\documentclass{article}
\usepackage{spconf,amsmath,graphicx}
\usepackage{multirow}
\usepackage[table,xcdraw]{xcolor}
\usepackage{afterpage}
\usepackage{fancyhdr}

\title{Breaking the Barrier: Selective Uncertainty-based Active Learning for Medical Image Segmentation}

\name{
    Siteng Ma\textsuperscript{1},
    Haochang Wu\textsuperscript{2},
    Aonghus Lawlor\textsuperscript{1},
    Ruihai Dong\textsuperscript{1},
}

\address{
    \textsuperscript{1}The Insight Centre for Data Analytics, School of Computer Science, University College Dublin, Dublin, Ireland\\
    \textsuperscript{2}School of Electrical and Electronic Engineering, University College Dublin, Dublin, Ireland
}

\begin{document}
%\ninept
%
\maketitle

\thispagestyle{fancy}
\fancyhead{}
\lhead{}
\lfoot{}
\cfoot{\small{© 2024 IEEE. Personal use of this material is permitted. Permission from IEEE must be obtained for all other uses, in any current or future media, including reprinting/republishing this material for advertising or promotional purposes, creating new collective works, for resale or redistribution to servers or lists, or reuse of any copyrighted component of this work in other works.}}
\rfoot{}

\begin{abstract}
Active learning (AL) has found wide applications in medical image segmentation, aiming to alleviate the annotation workload and enhance performance. Conventional uncertainty-based AL methods, such as entropy and Bayesian, often rely on an aggregate of all pixel-level metrics. However, in imbalanced settings, these methods tend to neglect the significance of target regions, eg., lesions, and tumors. Moreover, uncertainty-based selection introduces redundancy. These factors lead to unsatisfactory performance, and in many cases, even underperform random sampling. To solve this problem, we introduce a novel approach called the Selective Uncertainty-based AL, avoiding the conventional practice of summing up the metrics of all pixels. Through a filtering process, our strategy prioritizes pixels within target areas and those near decision boundaries. This resolves the aforementioned disregard for target areas and redundancy. Our method showed substantial improvements across five different uncertainty-based methods and two distinct datasets, utilizing fewer labeled data to reach the supervised baseline and consistently achieving the highest overall performance. Our code is available at https://github.com/HelenMa9998/Selective\_Uncertainty\_AL.
\end{abstract}

\begin{keywords}
Active learning, Uncertainty-based query strategy, Medical image segmentation
\end{keywords}

\section{Introduction}
\label{sec:intro}
Deep learning (DL) techniques have yielded great achievements in organ and abnormality segmentation \cite{litjens2017survey}. However, the training of models heavily relies on extensive datasets with pixel-level annotations, which is expert-oriented and costly \cite{shin2015interleaved,madani2018fast}. This dependence impedes the widespread adoption of DL in medical image diagnostics \cite{lee2017deep, campanella2019clinical}. Fortunately, Active Learning (AL) \cite{settles2009active} offers a potential solution by iteratively selecting informative samples for manual annotation, thus enhancing model performance while reducing the labeling workload.

In the existing research, numerous query strategies have emerged for sample selection. Uncertainty-based methods \cite{settles2009active,houlsby2011bayesian} are the most popular because of their simplicity \cite{ren2021survey}. They select samples based on the ambiguities of the model's output but tend to disregard the intrinsic data distribution. Furthermore, in the context of segmentation, contemporary AL methods continue to employ conventional classification-oriented techniques, such as entropy and Bayesian, through cumulative pixel-level metric aggregation \cite{gorriz2017cost,hiasa2019automated,shen2021labeling}. However, despite numerous studies showing that uncertainty-based methods underperform random sampling \cite{burmeister2022less, zhao2021dsal, shen2021labeling}, the core reasons behind this phenomenon remain largely unexplored.

In this study, we formulate hypotheses that investigate the reasons behind the suboptimal performance observed in current uncertainty-based methods, including the issue of high redundancy \cite{wang2018uncertainty, wu2022entropy} and the practice of uniformly aggregating metrics from all pixels. In response, we have developed a novel uncertainty-driven approach that seamlessly adapts to all existing uncertainty-based methods by selecting pivotal pixels to assess uncertainty indicators. One subset is derived from pixels within potential target regions, emphasizing these areas and alleviating the imbalanced problem in medical image segmentation tasks. To further mitigate redundancy, another subset leverages aggregated uncertainty metrics derived from pixels near decision boundaries for sample selection. By integrating these two subsets, we consistently achieve exceptional performance.

\section{Proposed Method}

\begin{figure}[t]
\centering
\includegraphics[width=0.34\textwidth]{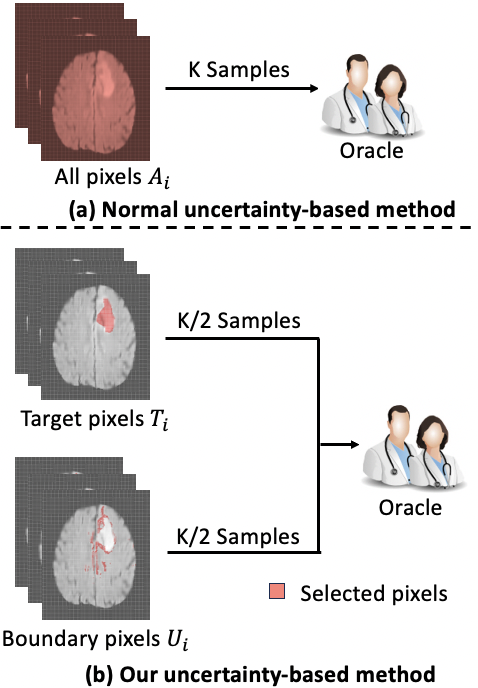}
\caption{\label{fig:method}Comparison between the conventional and our proposed methods. (a) The general framework of the conventional uncertainty-based query strategy; (b) Our proposed Selective Uncertainty-based method. }

% Comparison between the conventional and our proposed methods. (a) The general framework of the uncertainty-based query strategy in the segmentation task: aggregates uncertainty metrics of all pixels to select k samples. (b) Our Selective Uncertainty-based Selection. We choose the most informative pixels: target pixels and those near the decision boundary to obtain sample-level uncertainty for selection. To maintain diversity, we individually select k/2 from each category for combination.
\end{figure}

\subsection{Proposed Method}
In this section, we propose a selective uncertainty-based strategy that calculates the uncertainty index by filtering out significant pixels, rather than the conventional approach of aggregating metrics from all pixels. The comparison can be seen in Figure \ref{fig:method}: our strategy employs a multifaceted approach that leads to a more precise assessment of uncertainty, including both target-aware and boundary-driven selection. 

\subsubsection{Target-Aware Uncertainty Sampling}
Current uncertainty-based methods typically sum up pixel-level metrics uniformly, assuming that this adequately represents a sample. However, pixels exhibit varying degrees of importance. Uniform metric summation may cause valuable contributions from target regions obscured by less relevant areas. This is especially critical in medical segmentation tasks, where small lesions may be overwhelmed by a large background region in the overall uncertainty distribution, thereby compromising the efficacy of sample selection. 

Therefore, we propose a solution that calculates the uncertainty index by selectively considering significant pixels over a comprehensive pixel-level aggregation. To concentrate on relevant target regions, such as areas with anomalies or regions of interest (e.g., an organ) for segmentation, we identify these target pixels based on a threshold-based criterion using the model's output. Specifically, for each sample $x$, predicted target pixels $T_{x}$ can be obtained by: 

\begin{equation}
T_{x} = \left \{ p|p\in pixels_{x},P(c|p)>T  \right \}
\end{equation}
where $pixels_{x}$ represent all the pixels in a sample $x$, $T$ is the threshold of the target area, and $P(c|p)$ denotes the probability that a pixel belongs to class c given the pixel $p$. 

\begin{table*}[ht]
\caption{Performance of Ablation Study. }
\label{tab1}
\centering
\resizebox{0.76\textwidth}{!}{%
\begin{tabular}{lrrrrrrrl}
\hline
Query&\multicolumn{6}{c}{Number of labeled data (\%)}& Highest & Number of Labeled Data\\
    \cline{2-7} 
methods & 10\%& 15\%&20\%& 25\%& 30\%& 35\% & \multicolumn{1}{c}{Dice}  & for supervised baseline\\
\hline
Entropy                        & 0.6622          & 0.6900          & 0.7409          & 0.7328          & \textbf{0.7499}                & \textbf{0.7529} & \multicolumn{1}{c}{0.7603}                                    & \multicolumn{1}{c}{/}                                                                \\
Entropy + target               & \textbf{0.6670} & \textbf{0.7542} & \textbf{0.7465} & \textbf{0.7519} & \textbf{0.7615}                & 0.7300          & \multicolumn{1}{c}{\textbf{0.7732}}                                    & \multicolumn{1}{c}{\textbf{2100}}                                \\
Entropy + target + boundary    & \textbf{0.7318} & \textbf{0.7614} & \textbf{0.7779} & \textbf{0.7601} & \cellcolor[HTML]{FFFFFF}0.7290 & \textbf{0.7743} & \multicolumn{1}{c}{\textbf{0.7849}}                                    & \multicolumn{1}{c}{\textbf{1000}}                                \\ \hline
\end{tabular}}
\end{table*}

\subsubsection{Boundary-Driven Uncertainty Sampling}
% Building on the concept of uncertainty-driven sample selection discussed in the previous section, we now introduce an additional dimension to our uncertainty query strategy, called boundary-driven perspective, which complements our target-aware uncertainty sampling. 

Previous studies \cite{wang2018uncertainty,wu2022entropy} have shown that these methods tend to favor certain sample categories, consequently disregarding a wider range of valuable samples and leading to redundancy issues. This stems from the nature of uncertainty-based methods, which heavily rely on model predictions for selection. The model tends to produce consistent predictions for related samples, leading to a convergence of uncertainty metrics, resulting in redundant selections that hinder model optimization. Due to this, our uncertainty query strategy is further enriched by incorporating a boundary-centric perspective. Inherently, pixels close to the decision boundaries possess characteristics that straddle multiple classifications. By considering the uncertainty indices of these pixels, we can gain insight into the model's uncertainty near the classification boundary. Given $B$ represents the threshold to the decision boundary and $U$ is the decision boundary (0.5), the pixels near the decision boundary $U_{x}$ for a sample $x$ can be expressed as: 
\begin{equation}
U_{x} = \left \{ p|p\in pixels_{x},|P(c|p)-B|<U  \right \} 
\end{equation}

\subsubsection{Selective Uncertainty Sampling}
In this final phase, to ensure diversity selection, we merge the selections from both uncertainty exploration streams. Uncertainty-based indices are computed for qualifying pixels, providing uncertainty metrics for both the target region and the surrounding decision boundary area. This innovative approach mitigates the risk of introducing excessive redundancy in selected samples during each iteration, enabling a more comprehensive uncertainty exploration. For each sample $i$ in the dataset, the final selection can be defined as: 
\begin{equation}
Uncertainty\_T_{i} =  {\textstyle \sum_{p\in T_{i} }^{}Uncertainty\_Method(p)}
\end{equation}
\begin{equation}
Uncertainty\_U_{i} =  {\textstyle \sum_{p\in U_{i} }^{}Uncertainty\_Method(p)} 
\end{equation}
\begin{equation}
\begin{split}
Query &= TopK(Uncertainty\_T,k/2) \cup \\
&\quad TopK(Uncertainty\_U,k/2)
\end{split}
\end{equation}

Where TopK(X, k) selects the indices of the top k elements with the highest uncertainty from list X, enabling the joint selection of 2k samples per round based on both target and boundary uncertainty. This ensures diversity and effective utilization of information.

In conclusion, our method ensures a balanced representation of samples from target-aware and boundary-driven uncertainty sampling, encapsulating the advantages of precise task relevance and comprehensive uncertainty exploration.

\begin{table*}[ht]
\caption{Examination result for BraTS dataset. }
\label{tab2}
\centering
\resizebox{0.76\textwidth}{!}{%
\begin{tabular}{lrrrrrrrc}
\hline
Query&\multicolumn{6}{c}{Number of labeled data (\%)}& Highest & Number of Labeled Data\\
    \cline{2-7} 
Methods & 10\%& 15\%&20\%& 25\%& 30\%& 35\% & \multicolumn{1}{c}{Dice}& for supervised baseline \\
\hline
Random                                                           & 0.7035          & 0.7321          & 0.7262          & 0.7304          & 0.7462                         & 0.7631          & 0.7698                                             & /                                                                                               \\ \hline
Entropy                                                          & 0.6622          & 0.6900          & 0.7409          & 0.7328          & \textbf{0.7499}                & 0.7529          & 0.7603                                             & /                                                                                               \\
Entropy\_our                                                     & \textbf{0.7318} & \textbf{0.7614} & \textbf{0.7779} & \textbf{0.7601} & \cellcolor[HTML]{FFFFFF}0.7290 & \textbf{0.7743} & \textbf{0.7849}                                    & \textbf{1000 (12.9\%)}                                                                                   \\ \hline
LeastConfidence      & 0.7274          & \textbf{0.7580} & 0.7270          & 0.7430          & 0.7451                         & 0.7646          & 0.7646                                             & /                                                                                               \\
LeastConfidence\_our & \textbf{0.7563} & 0.7384          & \textbf{0.7534} & \textbf{0.7471} & \textbf{0.7452}                & \textbf{0.7669} & \textbf{0.7669}                                    & /                                                                                               \\ \hline
MarginSampling                               & 0.6729          & 0.6825          & 0.7322          & 0.6971          & 0.7145                         & 0.7346          & 0.7346                                             & /                                                                                               \\
MarginSampling\_our                          & \textbf{0.7332} & \textbf{0.7392} & \textbf{0.7454} & \textbf{0.7838} & \textbf{0.7445}                & \textbf{0.7838} & \textbf{0.7838}                                    & \textbf{1400 (18.1\%)}                                                                                   \\ \hline
MC-dropout                                                       & 0.7086          & 0.7255          & 0.7230          & 0.7525          & 0.7471                         & 0.7525          & 0.7525                                             & /                                                                                               \\
MC-dropout\_our                                                  & \textbf{0.7513} & \textbf{0.7597} & \textbf{0.7440} & \textbf{0.7773} & \textbf{0.7545}                & \textbf{0.7773} & \textbf{0.7773}                                    & \textbf{1900 (25.3\%)}                                                                                   \\ \hline
BALD                                                             & 0.7299          & 0.7073          & \textbf{0.7667} & 0.7480          & 0.7307                         & 0.7667          & 0.7667                                             & /                                                                                               \\
BALD\_our                                                        & \textbf{0.7361} & \textbf{0.7567} & 0.7107          & \textbf{0.7552} & \textbf{0.7712}                & \textbf{0.7735} & \textbf{0.7735}                                    & \textbf{2200 (28.34\%)}                                                                                   \\ \hline
\end{tabular}}
\end{table*}

\begin{table*}[ht]
\caption{Examination result for MSD dataset. }
\label{tab3}
\centering
\resizebox{0.76\textwidth}{!}{%
\begin{tabular}{lrrrrrrrl}
\hline

Query&\multicolumn{6}{c}{Number of labeled data (\%)}& Highest & Number of Labeled Data\\
    \cline{2-7} 
Methods & 10\%& 15\%&20\%& 25\%& 30\%& 35\% & \multicolumn{1}{c}{Dice}& for supervised baseline \\
\hline
    
Random                                                           & 0.8263          & 0.8865          & 0.8915          & 0.9273                   & 0.9147          & 0.9341          & \multicolumn{1}{c}{0.9278}                                             & \multicolumn{1}{c}{/}                                                                           \\ \hline
Entropy                                                          & 0.6663          & 0.6982          & 0.9062          & 0.8934                   & 0.9240          & 0.8926          & \multicolumn{1}{c}{0.9240}                                             & \multicolumn{1}{c}{/}                                                                           \\
Entropy\_our                                                     & \textbf{0.8305} & \textbf{0.9011} & \textbf{0.9216} & \textbf{0.9109} & \textbf{0.9475} & \textbf{0.9468} & \multicolumn{1}{c}{\textbf{0.9491}}                                    & \multicolumn{1}{c}{\textbf{700 (31.7\%)}}                                            \\ \hline
LeastConfidence      & 0.7709          & 0.7863          & 0.8006          & 0.8280                   & 0.8357          & 0.8831          & \multicolumn{1}{c}{0.9103}                                             & \multicolumn{1}{c}{/}                                                                           \\
LeastConfidence\_our & \textbf{0.8719} & \textbf{0.8906} & \textbf{0.9298} & \textbf{0.9398}          & \textbf{0.9365} & \textbf{0.9381} & \multicolumn{1}{c}{\textbf{0.9398}}                                    & \multicolumn{1}{c}{\textbf{550 (24.9\%)}}                                            \\ \hline
MarginSampling                              & 0.6565          & 0.8311          & 0.8643          & 0.8836                   & 0.9197          & 0.9429          & \multicolumn{1}{c}{0.9429}                                             & \multicolumn{1}{c}{800 (36.2\%)}                                                     \\
MarginSampling\_our                          & \textbf{0.8098} & \textbf{0.8405} & \textbf{0.9111} & \textbf{0.9286}          & \textbf{0.9327} & \textbf{0.9432} & \multicolumn{1}{c}{\textbf{0.9432}}                                    & \multicolumn{1}{c}{\textbf{750 (34.0\%)}}                                            \\ \hline
MC-dropout                                                       & 0.8118          & \textbf{0.8665} & 0.8897          & 0.9008                   & 0.9245          & \textbf{0.9361} & \multicolumn{1}{c}{0.9359}                                             & \multicolumn{1}{c}{/}                                                                           \\
MC-dropout\_our                                                  & \textbf{0.8246} & 0.8419          & \textbf{0.8977} & \textbf{0.9123}          & \textbf{0.9369} & 0.9329          & \multicolumn{1}{c}{\textbf{0.9369}}                                    & \multicolumn{1}{c}{/}                                                                           \\ \hline
BALD                                                             & 0.7727          & 0.8379          & 0.9009          & 0.9076                   & \textbf{0.9356} & 0.9245          & \multicolumn{1}{c}{0.9491}                                             & \multicolumn{1}{c}{\textbf{650 (29.4\%)}}                                            \\
BALD\_our                                                        & \textbf{0.8320} & \textbf{0.8673} & \textbf{0.9074} & \textbf{0.9097}          & 0.9310          & \textbf{0.9387} & \multicolumn{1}{c}{\textbf{0.9517}}                                    & \multicolumn{1}{c}{800 (36.2\%)}                                                     \\ \hline
\end{tabular}}
\end{table*}

\section{EXPERIMENT}
\subsection{Implementation Details}
\noindent\textbf{Dataset Description}
To show the efficacy of our method across various tasks, we selected diverse medical image segmentation tasks: BraTS Dataset \cite{bakas2017advancing} comprises 335 patients, including 259 high-grade gliomas (HGG) and 76 low-grade gliomas (LGG) with four modalities (T1, T2, T1ce, and Flair). LGG Flair is used to enhance tumor core segmentation in this study; Medical Segmentation Decathlon Dataset (MSD) \cite{simpson2019large} is an open challenge for segmentation. Spleen segmentation is one task provided, containing 61 CT studies.

\noindent\textbf{Data Processing} Given that 3D medical images are often annotated slice by slice, we adopted 2D slices for queries. The initial data size for BraTS and MSD is 500 and 150 respectively, with 100 and 50 queries per cycle. Online augmentation techniques, including Gaussian blur, rotations, and flips, were applied.

\begin{figure*}[h]
\centering
\includegraphics[width=0.76\textwidth]{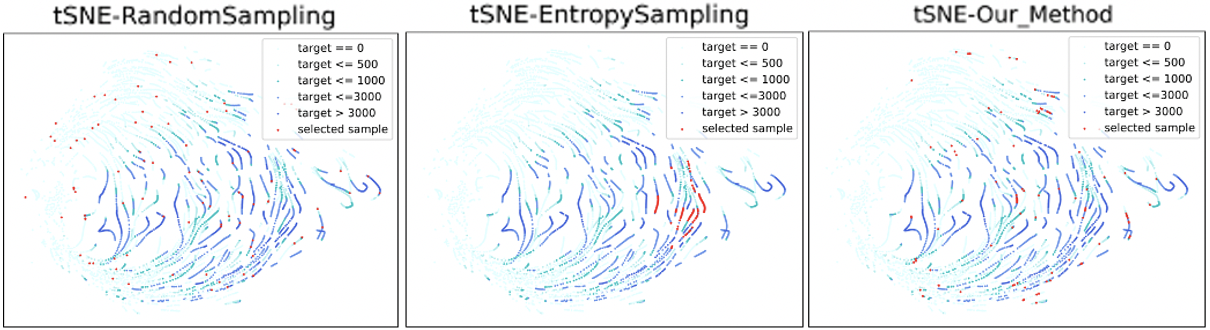}
\caption{\label{fig:distribution}Distribution of selected samples in the unlabeled pool: Random sampling selects diverse samples, but with many blank slices. The existing entropy-based method selects samples with targets but introduces redundancy. Our modified entropy-based method considers both the model's inherent prediction uncertainty and sample distribution. }
\end{figure*}

\noindent\textbf{Model and Hyper Parameters} We employed a designed Unet in Pytorch, with BCE loss. For a fair comparison, all experiments were done under NVIDIA GeForce RTX 4090 and consistent hyper-parameter settings were maintained across all experiments. The Adam optimizer was utilized with a learning rate of 0.0001. Batch sizes of 32 and 16 were chosen for BraTS and MSD, respectively, with a maximum epoch of 100. During each AL iteration, model training continued until validation loss became consistent. The parameter of the model is fine-tuned based on the previous round to speed up the process. 

\noindent\textbf{Evaluation Matrix}
The Dice coefficient, an ensemble similarity metric, was employed for the evaluation of segmentation. To evaluate the AL process, our analysis includes two aspects: the minimum number of manually labeled data required to achieve supervised learning performance and the peak performance achieved throughout the process.

\subsection{Ablation Studies}
To verify the validity of each module, we conducted ablation studies on BraTS, using entropy sampling as an example. The experimental results are summarized in Table \ref{tab1}. Notably, the symbol "/" indicates that the supervised baseline was not attained during this process. We first introduced target-aware sampling. The outcomes demonstrate that by focusing on the target regions, our approach consistently achieves improvements across nearly all training stages. It is worth highlighting that we reached the supervised performance with only 2100 samples and peaks at 0.7732, surpassing the original entropy by over 0.01. On this basis, we explored boundary-based selection: the subsequent enhancements are consistently above 0.01 compared to the sorely target-based method and utilize a minimal number of samples (1000), achieving the highest performance of 0.7849. We attribute this success to the diversity selection inherent in this module, which will be demonstrated in the Section.\ref{Experimental Results}.

% \subsection{Visualization}
% \label{Visualization}
% As a way of validating our hypothesis concerning diversity, we employ t-SNE to visualize the selection distribution of samples throughout the process. We compare the results of random, entropy (a traditional uncertainty-based method), and our modified version of entropy sampling. Fig. \ref{fig:distribution} displays these results, with shades of blue indicating the number of target-containing pixels in the patches: lighter shades for fewer targets and darker for more. Evidently, random sampling exhibits broad randomness, capturing diverse distributions but with many empty samples. Conventional entropy selection mostly picks samples that are rich in targets, yet introduce redundancy and miss the overall distribution, while our modified entropy-based method combines the advantages of diversity in random sampling, and addresses the model's inherent uncertainty that selects a substantial number of samples containing targets, thereby demonstrating the effectiveness of our method in sample selection. 

\subsection{Experimental Results}
\label{Experimental Results}
Aiming to demonstrate the generality of our approach, we used random sampling as the primary baseline and conducted experiments on five widely recognized uncertainty-based methods: LeastConfidence \cite{wang2016cost}, selecting samples with the lowest confidence; MarginSampling \cite{scheffer2001active}, choosing samples with the greatest prediction discrepancies; Entropy-based method \cite{settles2009active}, which picks samples with maximum prediction entropy; MC-Dropout \cite{houlsby2011bayesian}, employing multiple forward passes to estimate sample uncertainty; and Bayesian AL \cite{houlsby2011bayesian}, based on the mutual information between predictions and the model posterior.

Detailed experimental results for the BraTS dataset can be seen in Table \ref{tab2}. It can be observed that random selection exhibits a significant advantage over classical uncertainty-based methods, especially in the early stages, validating the motivation behind this paper. Moreover, our proposed method demonstrates adaptability to the majority of existing uncertainty-based methods, consistently outperforming them in most rounds, with an average improvement of 0.02. Importantly, while original uncertainty-based methods consistently fail to reach the supervised baseline (0.7714) throughout the process, our method accomplishes this with fewer samples (1000, 1400, 1900, and 2200) and achieves the highest dice score. For example, in Margin Sampling, our method eventually reaches a value of 0.7838, compared to the original version's 0.7346. Additionally, our approach consistently maintains notably better performance than random sampling, i.e., outperforming it by approximately 0.02 or more in most cases, demonstrating the effective resolution of the limited performance of existing uncertainty-based methods.

Similar performance can be observed in MSD, as shown in Table \ref{tab3}. Our approach consistently outperforms the baselines (over 0.05) per round. We achieved supervised baseline performance (0.9376) with fewer labeled samples (700, 550, 750), resulting in overall performance improvements. Moreover, our approach demonstrates superior performance compared to random selection in the majority of cases.

To validate our hypothesis regarding diversity, we employ t-SNE to visualize the selection distribution of samples throughout the process. We compare the results of random, conventional entropy (a traditional uncertainty-based method), and our modified version of entropy sampling. Figure \ref{fig:distribution} displays these results, with shades of blue indicating the number of target-containing pixels in the slices: lighter shades for fewer targets and darker for more. Evidently, random sampling exhibits broad randomness, capturing diverse distributions but with many empty samples. Conventional entropy selection mostly picks samples that are rich in targets, yet introduce redundancy and miss the overall distribution, while our modified entropy-based method combines the advantages of diversity in random sampling, and addresses the model's inherent uncertainty that selects a substantial number of samples containing targets, thereby demonstrating the effectiveness of our method in sample selection. 

\section{Conclusions}
This paper addresses the limitations of uncertainty-based AL methods in medical image segmentation. Through theoretical analysis, we highlight challenges stemming from sample redundancy and uniform aggregation of all pixel-level metrics. Hence, we introduce a selective uncertainty-based query strategy, leveraging target-aware and boundary-driven uncertainty sampling. Our strategy not only enhances segmentation performance but also reduces the burden on domain experts. 

\section{Acknowledgement}
This research was conducted with the financial support of Science Foundation Ireland [12/RC/2289\_P2] at Insight the SFI Research Centre for Data Analytics at University College Dublin.

% \vfill\pagebreak

% \section{REFERENCES}
% \label{sec:refs}

\bibliographystyle{IEEEbib}
\bibliography{refs}

\begin{thebibliography}{10}

\bibitem{litjens2017survey}
Geert Litjens, Thijs Kooi, Babak~Ehteshami Bejnordi, Arnaud Arindra~Adiyoso
  Setio, Francesco Ciompi, Mohsen Ghafoorian, Jeroen~Awm Van Der~Laak, Bram
  Van~Ginneken, and Clara~I S{\'a}nchez,
\newblock ``A survey on deep learning in medical image analysis,''
\newblock {\em Medical image analysis}, vol. 42, pp. 60--88, 2017.

\bibitem{shin2015interleaved}
Hoo-Chang Shin, Le~Lu, Lauren Kim, Ari Seff, Jianhua Yao, and Ronald~M Summers,
\newblock ``Interleaved text/image deep mining on a very large-scale radiology
  database,''
\newblock in {\em Proceedings of the IEEE conference on computer vision and
  pattern recognition}, 2015, pp. 1090--1099.

\bibitem{madani2018fast}
Ali Madani, Ramy Arnaout, Mohammad Mofrad, and Rima Arnaout,
\newblock ``Fast and accurate view classification of echocardiograms using deep
  learning,''
\newblock {\em NPJ digital medicine}, vol. 1, no. 1, pp. 6, 2018.

\bibitem{lee2017deep}
June-Goo Lee, Sanghoon Jun, Young-Won Cho, Hyunna Lee, Guk~Bae Kim, Joon~Beom
  Seo, and Namkug Kim,
\newblock ``Deep learning in medical imaging: general overview,''
\newblock {\em Korean journal of radiology}, vol. 18, no. 4, pp. 570--584,
  2017.

\bibitem{campanella2019clinical}
Gabriele Campanella, Matthew~G Hanna, Luke Geneslaw, Allen Miraflor, Vitor
  Werneck Krauss~Silva, Klaus~J Busam, Edi Brogi, Victor~E Reuter, David~S
  Klimstra, and Thomas~J Fuchs,
\newblock ``Clinical-grade computational pathology using weakly supervised deep
  learning on whole slide images,''
\newblock {\em Nature medicine}, vol. 25, no. 8, pp. 1301--1309, 2019.

\bibitem{settles2009active}
Burr Settles,
\newblock ``Active learning literature survey,''
\newblock 2009.

\bibitem{houlsby2011bayesian}
Neil Houlsby, Ferenc Husz{\'a}r, Zoubin Ghahramani, and M{\'a}t{\'e} Lengyel,
\newblock ``Bayesian active learning for classification and preference
  learning,''
\newblock {\em arXiv preprint arXiv:1112.5745}, 2011.

\bibitem{ren2021survey}
Pengzhen Ren, Yun Xiao, Xiaojun Chang, Po-Yao Huang, Zhihui Li, Brij~B Gupta,
  Xiaojiang Chen, and Xin Wang,
\newblock ``A survey of deep active learning,''
\newblock {\em ACM computing surveys (CSUR)}, vol. 54, no. 9, pp. 1--40, 2021.

\bibitem{gorriz2017cost}
Marc Gorriz, Axel Carlier, Emmanuel Faure, and Xavier Giro-i Nieto,
\newblock ``Cost-effective active learning for melanoma segmentation,''
\newblock {\em arXiv preprint arXiv:1711.09168}, 2017.

\bibitem{hiasa2019automated}
Yuta Hiasa, Yoshito Otake, Masaki Takao, Takeshi Ogawa, Nobuhiko Sugano, and
  Yoshinobu Sato,
\newblock ``Automated muscle segmentation from clinical ct using bayesian u-net
  for personalized musculoskeletal modeling,''
\newblock {\em IEEE transactions on medical imaging}, vol. 39, no. 4, pp.
  1030--1040, 2019.

\bibitem{shen2021labeling}
Maohao Shen, Jacky~Y Zhang, Leihao Chen, Weiman Yan, Neel Jani, Brad Sutton,
  and Oluwasanmi Koyejo,
\newblock ``Labeling cost sensitive batch active learning for brain tumor
  segmentation,''
\newblock in {\em 2021 IEEE 18th International Symposium on Biomedical Imaging
  (ISBI)}. IEEE, 2021, pp. 1269--1273.

\bibitem{burmeister2022less}
Josafat-Mattias Burmeister, Marcel~Fernandez Rosas, Johannes Hagemann, Jonas
  Kordt, Jasper Blum, Simon Shabo, Benjamin Bergner, and Christoph Lippert,
\newblock ``Less is more: A comparison of active learning strategies for 3d
  medical image segmentation,''
\newblock {\em arXiv preprint arXiv:2207.00845}, 2022.

\bibitem{zhao2021dsal}
Ziyuan Zhao, Zeng Zeng, Kaixin Xu, Cen Chen, and Cuntai Guan,
\newblock ``Dsal: Deeply supervised active learning from strong and weak
  labelers for biomedical image segmentation,''
\newblock {\em IEEE journal of biomedical and health informatics}, vol. 25, no.
  10, pp. 3744--3751, 2021.

\bibitem{wang2018uncertainty}
Gaoang Wang, Jenq-Neng Hwang, Craig Rose, and Farron Wallace,
\newblock ``Uncertainty-based active learning via sparse modeling for image
  classification,''
\newblock {\em IEEE Transactions on Image Processing}, vol. 28, no. 1, pp.
  316--329, 2018.

\bibitem{wu2022entropy}
Jiaxi Wu, Jiaxin Chen, and Di~Huang,
\newblock ``Entropy-based active learning for object detection with progressive
  diversity constraint,''
\newblock in {\em Proceedings of the IEEE/CVF Conference on Computer Vision and
  Pattern Recognition}, 2022, pp. 9397--9406.

\bibitem{bakas2017advancing}
Spyridon Bakas, Hamed Akbari, Aristeidis Sotiras, Michel Bilello, Martin
  Rozycki, Justin~S Kirby, John~B Freymann, Keyvan Farahani, and Christos
  Davatzikos,
\newblock ``Advancing the cancer genome atlas glioma mri collections with
  expert segmentation labels and radiomic features,''
\newblock {\em Scientific data}, vol. 4, no. 1, pp. 1--13, 2017.

\bibitem{simpson2019large}
Amber~L Simpson, Michela Antonelli, Spyridon Bakas, Michel Bilello, Keyvan
  Farahani, Bram Van~Ginneken, Annette Kopp-Schneider, Bennett~A Landman, Geert
  Litjens, Bjoern Menze, et~al.,
\newblock ``A large annotated medical image dataset for the development and
  evaluation of segmentation algorithms,''
\newblock {\em arXiv preprint arXiv:1902.09063}, 2019.

\bibitem{wang2016cost}
Keze Wang, Dongyu Zhang, Ya~Li, Ruimao Zhang, and Liang Lin,
\newblock ``Cost-effective active learning for deep image classification,''
\newblock {\em IEEE Transactions on Circuits and Systems for Video Technology},
  vol. 27, no. 12, pp. 2591--2600, 2016.

\bibitem{scheffer2001active}
Tobias Scheffer, Christian Decomain, and Stefan Wrobel,
\newblock ``Active hidden markov models for information extraction,''
\newblock in {\em International symposium on intelligent data analysis}.
  Springer, 2001, pp. 309--318.

\end{thebibliography}

\end{document}